%% file: AnonymousSubmission2027.tex
\documentclass[letterpaper]{article} 
\usepackage[preprint]{aaai2027}  
\usepackage[hyphens]{url}  
\usepackage{graphicx} 
\usepackage{natbib}  
\usepackage{caption} 
\usepackage{amsmath}
\usepackage{amssymb}
\usepackage{booktabs}
\usepackage{makecell}
\usepackage{tabularx}
\usepackage{pifont}
\usepackage{xspace}
\usepackage{xcolor}
\usepackage{colortbl}
\usepackage{cleveref}
\usepackage{tikz}

\newcommand{\onedot}{.\xspace}
\newcommand{\eg}{\emph{e.g}\onedot}

\newcommand{\cmark}{\textcolor{green!70!black}{\ding{51}}}
\newcommand{\xmark}{\textcolor{red!70!black}{\ding{55}}}
\newcommand{\best}[1]{\textbf{#1}}
\newcommand{\second}[1]{\smash{\underline{#1}}\vphantom{#1}}
\newcommand{\oursrow}{}
\newcolumntype{Y}{>{\centering\arraybackslash}X}

\crefname{section}{Sec.}{Secs.}
\Crefname{section}{Section}{Sections}
\crefname{subsection}{Sec.}{Secs.}
\Crefname{subsection}{Section}{Sections}
\crefname{table}{Tab.}{Tabs.}
\Crefname{table}{Table}{Tables}

\title{TRACE: High-Fidelity 3D Scene Editing via Tangible Reconstruction and Geometry-Aligned Contextual Video Masking}

\author{
    Jiyuan Hu\textsuperscript{\rm 1},
    Zechuan Zhang\textsuperscript{\rm 1,\rm 2},
    Zongxin Yang\textsuperscript{\rm 1,\rm 2},
    Yi Yang\textsuperscript{\rm 1}
}
\affiliations{
    \textsuperscript{\rm 1}ReLER Lab, CCAI, Zhejiang University\\
    \textsuperscript{\rm 2}DBMI, HMS, Harvard University\\
}

\begin{document}

\maketitle

\input{sec/0_abstract}

\input{sec/1_intro}

\begin{figure*}[t]
    \centering
    \includegraphics[width=0.92\textwidth]{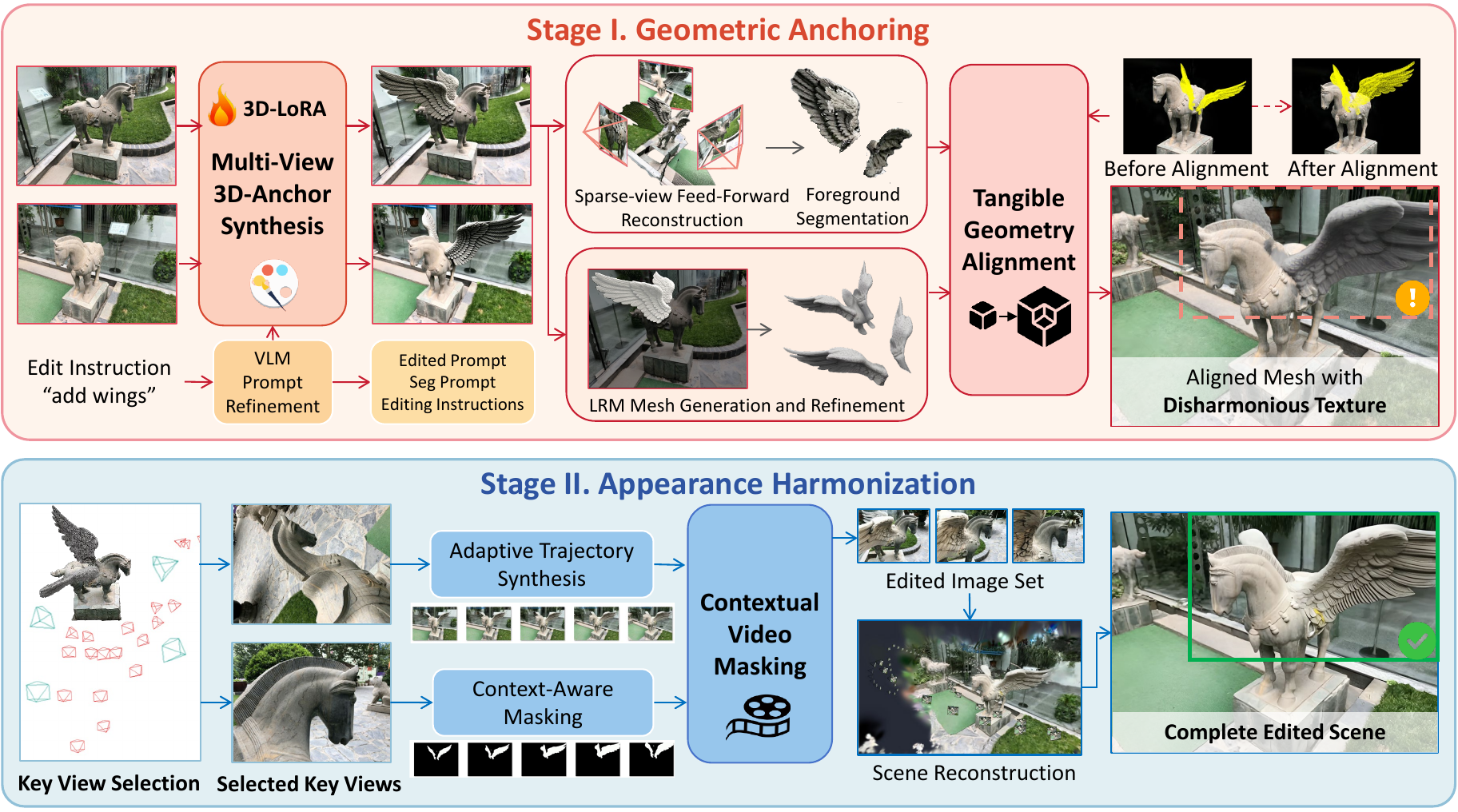}
    \caption{Overview of TRACE. Given a 3DGS scene and an editing instruction, Stage I generates sparse edited anchors and aligns a generated or retrieved mesh to the scene with the TGA module. Stage II projects the aligned mesh anchor along rendered camera trajectories and uses CVM to repaint the object and nearby context before reconstructing the edited 3DGS.}
    \label{fig:pipeline_overview}
\end{figure*}

\input{sec/2_related}

\begin{figure*}[t]
  \centering
  \includegraphics[width=\textwidth]{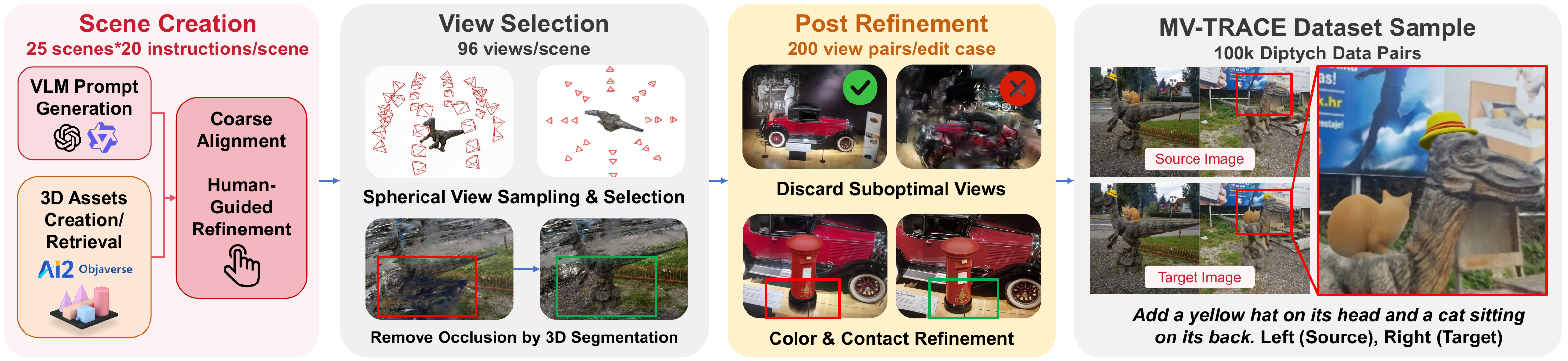}
  \caption{Construction of MV-TRACE. The 100k pairs come from 96-view rendering of 25 training scenes, automatic filtering, and human-assisted asset alignment and color/contact refinement; all construction scenes are disjoint from evaluation.}
  \label{fig:dataset_curation}
\end{figure*}

\input{sec/3_method}
\input{sec/4_experiment}
\input{sec/5_conclusion}

\bibliography{main}

\end{document}

%% file: sec/0_abstract.tex
\begin{abstract}
Existing 3D Gaussian Splatting (3DGS) editing methods primarily focus on appearance modification and often struggle to support flexible geometry editing while preserving structural integrity and scene-consistent appearance. To address this limitation, we present \textbf{TRACE}, a mesh-guided 3DGS editing framework that automatically aligns explicit 3D geometry with Gaussian scenes and decouples \textbf{Geometric Anchoring} from \textbf{Appearance Harmonization}. First, Multi-view 3D-Anchor Synthesis, trained on our \textbf{MV-TRACE} dataset for scene-coherent object addition and modification, generates geometrically aligned editing anchors, while Tangible Geometry Alignment (TGA) performs coarse-to-fine mesh--scene registration. Then, Contextual Video Masking (CVM) integrates projected 3D anchors into an autoregressive video diffusion pipeline, harmonizing their appearance with the surrounding scene while maintaining multi-view consistency. We evaluate TRACE on eight held-out scenes across six editing categories. TRACE completes each edit in $\sim$10 minutes on a single NVIDIA RTX Pro6000 GPU. Extensive experiments demonstrate consistent improvements over existing methods in editing versatility, structural integrity, semantic alignment, multi-view consistency, and visual quality.
\end{abstract}

%% file: sec/1_intro.tex
\section{Introduction}\label{sec:intro}
Recent advances in 3D Gaussian Splatting (3DGS)~\cite{kerbl3Dgaussians} have redefined high-fidelity real-time rendering, offering significant potential for applications ranging from interactive content creation to robotic simulation. However, achieving high-fidelity 3DGS scene editing with diverse functionalities---ranging from texture refinement to structural transformation---remains a significant challenge.

Existing 3D scene editing methods follow two primary paradigms based on their objectives. First, \textbf{appearance-centric frameworks}~\cite{zhao2025tinker, jiang2025vace, wang2025view} focus on global or local style and color manipulation. These methods anchor 2D generative priors with depth, epipolar, or rendered-view constraints to propagate edits, but they are not designed for object addition, shape deformation, or topology changes. Second, \textbf{structural modification methods}~\cite{chen2023gaussianeditor, wang2025personalize, 10887779} introduce explicit geometric proxies, such as meshes or bounding boxes, or rely on SDS-based optimization~\cite{poole2022dreamfusion, zhuang2024tip, Sun2024GSEditPro3G}. They improve structural control, but often require per-scene optimization, manual placement, or fragile alignment. As summarized in~\cref{tab:capability_comparison}, a practical 3DGS editor should jointly support \textbf{global/local style editing}, \textbf{partial/whole object geometry editing}, and \textbf{automatic 3D prior alignment}.

\begin{figure}[t]
  \centering
  \includegraphics[width=\columnwidth]{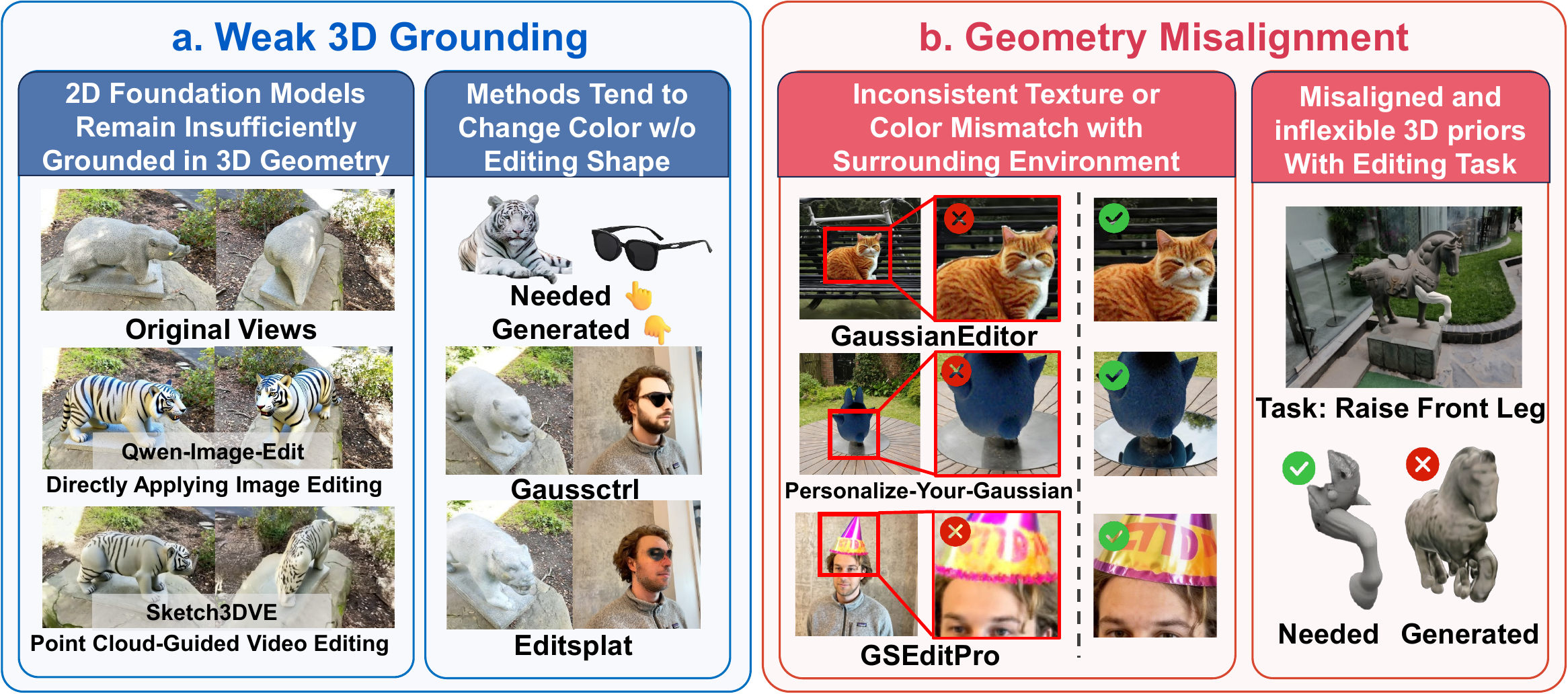}
  \caption{Limitations of existing 3D editing approaches. (a) Weak 3D grounding leads to inconsistent or appearance-only edits. (b) Explicit geometry suffers from visual mismatch and inflexible alignment.}
  \label{fig:2_difficulties}
\end{figure}

\begin{figure*}[t]
    \centering
    \includegraphics[width=0.92\textwidth]{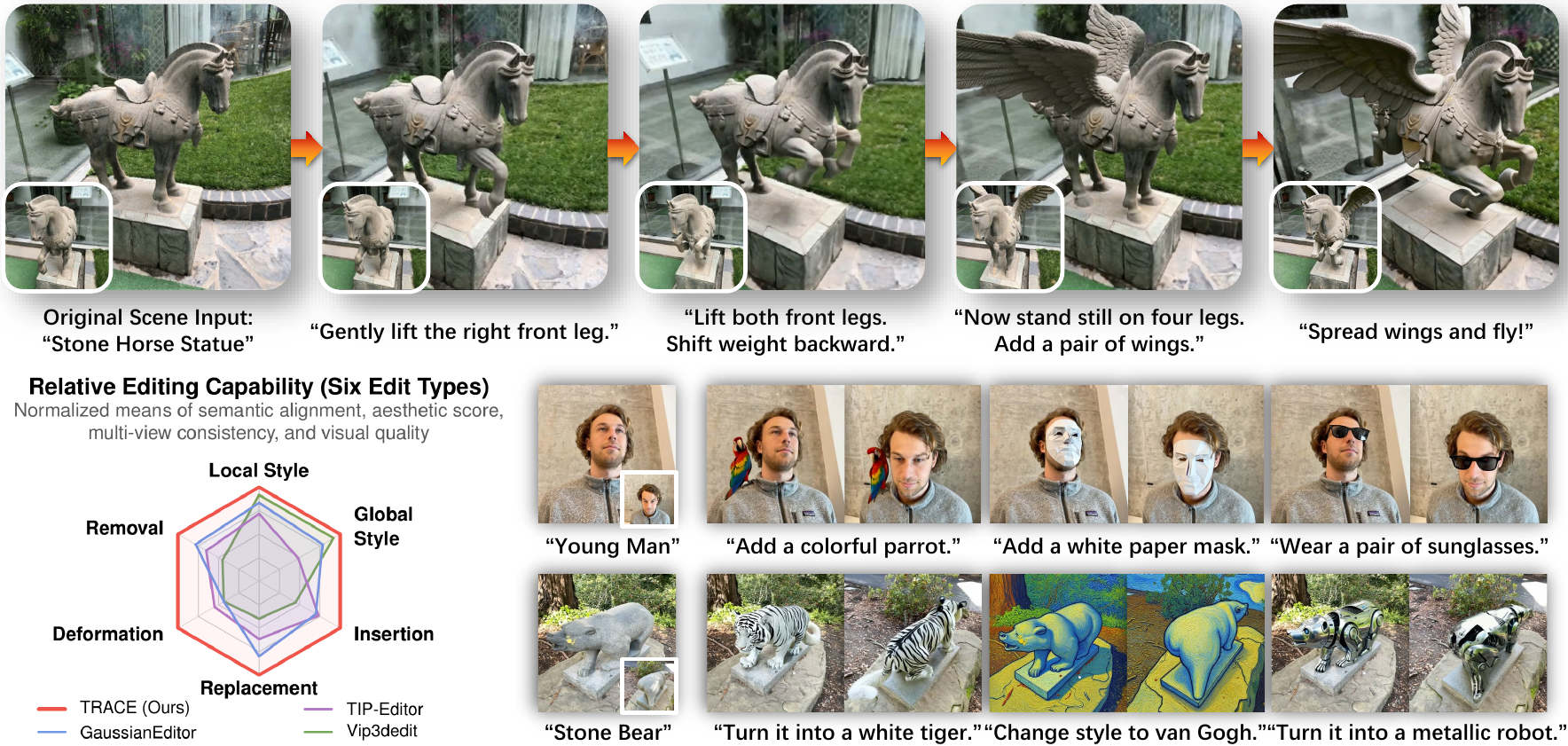}
    \caption{Representative TRACE edits. Top: sequential interactive edits on the ``Stone Horse'' scene, demonstrating part-level geometric manipulation. Bottom left: normalized performance across six edit types. Bottom right: multi-view examples of object addition, texture synthesis, style transfer, and replacement. Scenes come from IN2N, BlendedMVS, and Mip-NeRF 360.}
    \label{fig:teaser}
\end{figure*}

In spite of this progress, high-fidelity shape editing in 3DGS scenes remains challenging due to two limitations. \textbf{Geometric Instability due to Weak 3D Grounding.} Purely 2D-driven generative methods lack sufficient 3D grounding, leading to viewpoint ambiguity, back-side artifacts, and geometry drift in sparse-view editing (see Fig.~\ref{fig:2_difficulties}(a), left). Moreover, geometry--appearance entanglement can cause methods to mimic a requested geometry edit through color or texture changes without producing the intended shape modification (see Fig.~\ref{fig:2_difficulties}(a), right). \textbf{Inflexibility and Misalignment of Explicit Geometry.} Although explicit meshes provide spatial scaffolding for object addition and deformation, their raw appearance often conflicts with the target scene in texture, color, illumination, shadows, and boundaries (see Fig.~\ref{fig:2_difficulties}(b), left). Generated or retrieved meshes may be misaligned with the editing target in pose, scale, orientation, or local structure, while manual alignment requires extensive tuning and remains inflexible for fine-grained edits (see Fig.~\ref{fig:2_difficulties}(b), right). 

Our key observation is that explicit meshes and generative video priors solve complementary parts of the problem. A mesh supplies a stable 3D scaffold for insertion, deformation, and part-level pose changes, but its rendered appearance rarely matches the target scene. Video diffusion can harmonize texture, illumination, shadows, and boundaries across nearby views, but does not by itself enforce a desired 3D configuration. TRACE therefore treats the mesh as an editing anchor: geometry is aligned first, and its projection is then repainted together with the surrounding context.

Based on this strategy, we introduce \textbf{TRACE}, a two-stage 3DGS editing framework that decouples geometry anchoring from appearance harmonization. \textbf{Stage I: Geometric Anchoring} combines Multi-view 3D-Anchor Synthesis, trained on our 100k-pair MV-TRACE dataset, with Tangible Geometry Alignment (TGA) for automatic mesh-scene alignment. \textbf{Stage II: Appearance Harmonization} uses Contextual Video Masking (CVM) to repaint mesh projections, object boundaries, shadows, and surrounding context through video diffusion. Among existing methods, TRACE uniquely enables the three capabilities in~\cref{tab:capability_comparison}.

\begin{table}[t]
  \centering
  \footnotesize
  \setlength{\tabcolsep}{0.8pt}
  \begin{tabular*}{\columnwidth}{@{\extracolsep{\fill}}llccc@{}}
    \toprule
    \textbf{Method} & \textbf{Consistency} & \makecell[c]{\textbf{Shape} \\ \textbf{Edit}} & \makecell[c]{\textbf{Global} \\ \textbf{Edit}} & \makecell[c]{\textbf{Auto} \\ \textbf{Align}} \\
    \midrule
    DGE             & Extrap. Attn.   & \xmark & \cmark & --- \\
    GaussCtrl       & Depth Const.    & \xmark & \cmark & --- \\
    TIP-Editor      & SDS-based       & \cmark & \xmark & \xmark \\
    EditSplat       & Iter. Update    & \xmark & \cmark & --- \\
    GaussianEditor  & Iter. Update    & \cmark & \cmark & \xmark \\
    Vip3dedit       & Video/3D Prior  & \xmark & \cmark & --- \\
    Tinker          & Video Editing   & \xmark & \cmark & --- \\
    FreeInsert      & Mesh Prior      & \cmark & \xmark & \cmark \\
    \midrule
    \textbf{Ours}   & \textbf{Video/Mesh Prior} & \cmark & \cmark & \cmark \\
    \bottomrule
  \end{tabular*}
  \caption{Comparison of editing capabilities. Shape Edit means explicit 3D geometry modification; \cmark/\xmark denote claimed native support/unsupported, and "---" means Auto Align is inapplicable because no explicit asset is introduced.}
  \label{tab:capability_comparison}
\end{table}

Under an equal-weight average over the global/local style editing and partial/whole object geometry editing groups, TRACE reaches $\text{CLIP}_{\text{dir}}=\textbf{0.1514}$ and Aesthetic score $=\textbf{6.10}$ under the reported common-prompt protocol. Per-group scores are reported in~\cref{tab:quantitative_results}. TRACE completes a full edit in about \textbf{10 minutes} on one NVIDIA RTX Pro6000: $\sim$2 minutes for multi-view image editing, 3 minutes for mesh generation and TGA alignment, 3 minutes for masked video editing, and 2 minutes for final reconstruction. Representative edits and performance comparisons are shown in~\cref{fig:teaser}.

In summary, our main contributions are as follows:
\begin{itemize}
\item We introduce TRACE, a mesh-guided two-stage 3DGS editing pipeline that decouples Geometric Anchoring from Appearance Harmonization, enabling appearance editing, geometry editing, and automatic 3D-prior alignment in a unified pipeline while maintaining structural and scene-consistent appearance.
\item Motivated by the 3D inconsistency of off-the-shelf editors across sparse views, we construct MV-TRACE, to our knowledge the first multi-view consistent dataset for scene-coherent object addition and modification, and use it to train Multi-view 3D-Anchor Synthesis for robust sparse-view geometric grounding.
\item We introduce TGA for automatic mesh-scene alignment and CVM for appearance harmonization, turning explicit 3D anchors into seamless edits. Experiments across six edit types show stronger semantic alignment, multi-view consistency, visual quality, and editing efficiency.
\end{itemize}

%% file: sec/2_related.tex
\section{Related Work}
\subsection{2D Editing}
\noindent\textbf{i) Image Editing.}
Training-free editors~\cite{routSemanticImageInversion2025,esserScalingRectifiedFlow2024} use inversion or attention control for zero-shot edits~\cite{songDenoisingDiffusionImplicit2020,hertzPrompttoPromptImageEditing2022,zhang2025context}, while optimization/training-based editors improve controllability with paired data or test-time refinement~\cite{bar-talText2LIVETextDrivenLayered2022,yu2025anyedit}. Recent scaled image editors and autoregressive models further strengthen multimodal alignment~\cite{wu2025qwen,wu2025omnigen2}.

\noindent\textbf{ii) Video Editing.}
Video editing contributes sequence-level generative priors for rendered 3D scene views. Earlier methods extend T2I models to frame sequences~\cite{tokenflow2023,blattmann2023stable}, while recent editing systems~\cite{gu2023videoswap,dongWanAlphaHighQualityTexttoVideo2025,ku2024anyv2v} improve sequence-conditioned or training-free editing.

\subsection{3D Scene Editing}
\noindent\textbf{i) Iterative 2D-to-3D Editing.}
Instruct-NeRF2NeRF~\cite{instructnerf2023}, GaussianEditor, and GSEditPro propagate 2D edits into 3D scenes, but remain sensitive to editor stochasticity and view drift~\cite{zhang20243ditscene,kim2024dreamcatalyst}.

\noindent\textbf{ii) Personalized and Geometry-Constrained Editing.}
Personalization and geometry-constrained methods~\cite{chen2024shap,gomel2024diffusionbasedattentionwarpingconsistent}, such as TIP-Editor and Personalize Your Gaussian, use SDS, bounding boxes, DreamBooth, or LoRAs for content binding, but require costly per-scene optimization. Implicit alignment methods~\cite{zhu2026coreeditorcorrespondenceconstraineddiffusionconsistent}, including DGE and Tinker, rely on epipolar/video priors and still lack explicit, reusable geometry anchoring; Tinker primarily targets multi-view style transfer rather than scene-coherent localized geometry editing.

\noindent\textbf{iii) Native and Mesh-Guided Editing.}
VF-Editor~\cite{qin2026variationawareflexible3dgaussian} predicts Gaussian primitive variations but mainly supports semantic shifts. Mesh-guided and Image-to-3D methods improve structural control~\cite{chen2026sam,chang2025reconviagen,barda2024instant3ditmultiviewinpaintingfast}. Recent GeM-NR~\cite{bengtson2026gem} handles nonrigid multi-view edits, but its edited-view propagation can degrade under large viewpoint changes. Relative to sparse-view/video-prior editors such as Tinker~\cite{zhao2025tinker} and correspondence-constrained methods such as CoreEditor~\cite{zhu2026coreeditorcorrespondenceconstraineddiffusionconsistent}, TRACE makes the geometry path explicit: a registered mesh fixes the target configuration before CVM refinement.

%% file: sec/3_method.tex
\section{Method}
Given a reconstructed 3DGS scene $\mathcal{G}$, calibrated COLMAP/NeRF-format cameras $\mathcal{C}$, and an instruction $y$, our goal is an edited scene $\mathcal{G}^*$ that satisfies $y$ while preserving content outside the target region. TRACE first establishes a scene-aligned mesh anchor and then harmonizes its rendered projection with the surrounding scene (see~\cref{fig:pipeline_overview}).

\subsection{Stage I: Geometric Anchoring}
\subsubsection{Step 1: Multi-view 3D-Anchor Synthesis.}
\label{subsec:multiview_consistent_editing}
We render the 3DGS scene along a camera trajectory to obtain $\mathcal{V} = \{I_1, I_2, \dots, I_n\}$. Qwen-Image-Edit edits the first frame into the visual anchor $I^*_1$, while Qwen-VL handles prompt routing, spatial parsing, and contact-point descriptions. A common way to encourage multi-view consistency in off-the-shelf editors is to horizontally concatenate multiple views (\eg, triplets) and edit them jointly. However, these editors often oversimplify the task by placing the object at the same relative 2D canvas coordinates across views, rather than transforming its 3D pose with the changing camera. We therefore use iterative dual-view in-context editing: for each target view $I_t$, we concatenate $[I^*_1 \oplus I_t]$ and use VLM-derived contact points to form a spatial prompt $y_{iter}$ that constrains object-scene relations across views (see~\cref{fig:pipeline_overview}, top left).

\paragraph{Data Curation and Training Objective.} Existing multi-view datasets (\eg, MVImgNet~\cite{yu2023mvimgnet} and 360-USID~\cite{wu2025aurafusion360}) predominantly contain isolated objects or scene-level imagery, but lack scene-coherent annotations for localized addition or replacement. Tinker's dataset~\cite{zhao2025tinker} is limited to style transfer rather than geometry editing. To fill this gap, we fine-tune Qwen-Image-Edit with LoRA on MV-TRACE, a 100k-pair dataset built from 25 training scenes using 96 spherical views and human-aligned retrieved~\cite{objaverseXL} or generated assets (see~\cref{fig:dataset_curation}). Training and evaluation scenes, views, and edited pairs are disjoint. We prioritize view pairs with large horizontal and vertical angular disparity to learn 3D consistency; optimizer and LoRA details are in the supplement. For a training tuple $(I_{ref}^*, I_t, I_t^*, y_{iter})$, the edited anchor $I_{ref}^*$, target view $I_t$, and spatial prompt $y_{iter}$ form the image-editing condition $c$. Let $\mathcal{E}$ be the VAE encoder, $\mathbf{z}_0=\mathcal{E}(I_t^*)$ the edited-target latent, $\mathbf{z}_1\sim\mathcal{N}(0,\mathbf{I})$ a noise latent, and $\gamma\sim\mathcal{U}(0,1)$ the flow-matching time. We train the LoRA parameters $\theta$ with the latent flow-matching objective:
\begin{equation}
\begin{aligned}
\min_{\theta}\;&
\mathbb{E}_{\mathbf{z}_0,\mathbf{z}_1,\gamma,c}
\left[
\left\|v_{\theta}(\mathbf{z}_{\gamma},\gamma,c)
-(\mathbf{z}_0-\mathbf{z}_1)\right\|_2^2
\right],\\
&\mathbf{z}_{\gamma}=(1-\gamma)\mathbf{z}_1+\gamma\mathbf{z}_0 .
\end{aligned}
\end{equation}
Here, $v_{\theta}$ predicts the rectified-flow velocity from the interpolated latent $\mathbf{z}_{\gamma}$ to the target latent under condition $c$.

\subsubsection{Step 2: Tangible Geometry Alignment (TGA).}
\label{subsec:3d_prior_construction}
To bridge 2D edits with 3D space, we generate 3D priors and align them to multi-view masks in the original scene. Existing objects use original point clouds; new objects use sparse-view feed-forward reconstruction for localization. Coarse alignment renders six canonical mesh views, initializes orientation, and checks correspondences before differentiable registration.

\paragraph{Task-Adaptive 3D Prior Generation.}
To accommodate diverse editing instructions, we adopt a task-specific asset generation strategy. Qwen-VL is used for prompt routing and contact-point descriptions in anchor synthesis, while VLM-FO1~\cite{liu2025vlm} provides segmentation-oriented grounding prompts and SAM3 provides 2D masks for TGA. Replacement and deformation follow a remove-and-insert scheme at the whole-object and part levels, respectively, with the new geometry aligned by TGA. SAM 3D Objects~\cite{chen2026sam} is used for mesh generation, while style edits bypass TGA and use masked local or full-frame reconstruction for local and global editing (see~\cref{fig:pipeline_overview}, top middle).

\begin{figure}[t]
    \centering
    \includegraphics[width=\columnwidth]{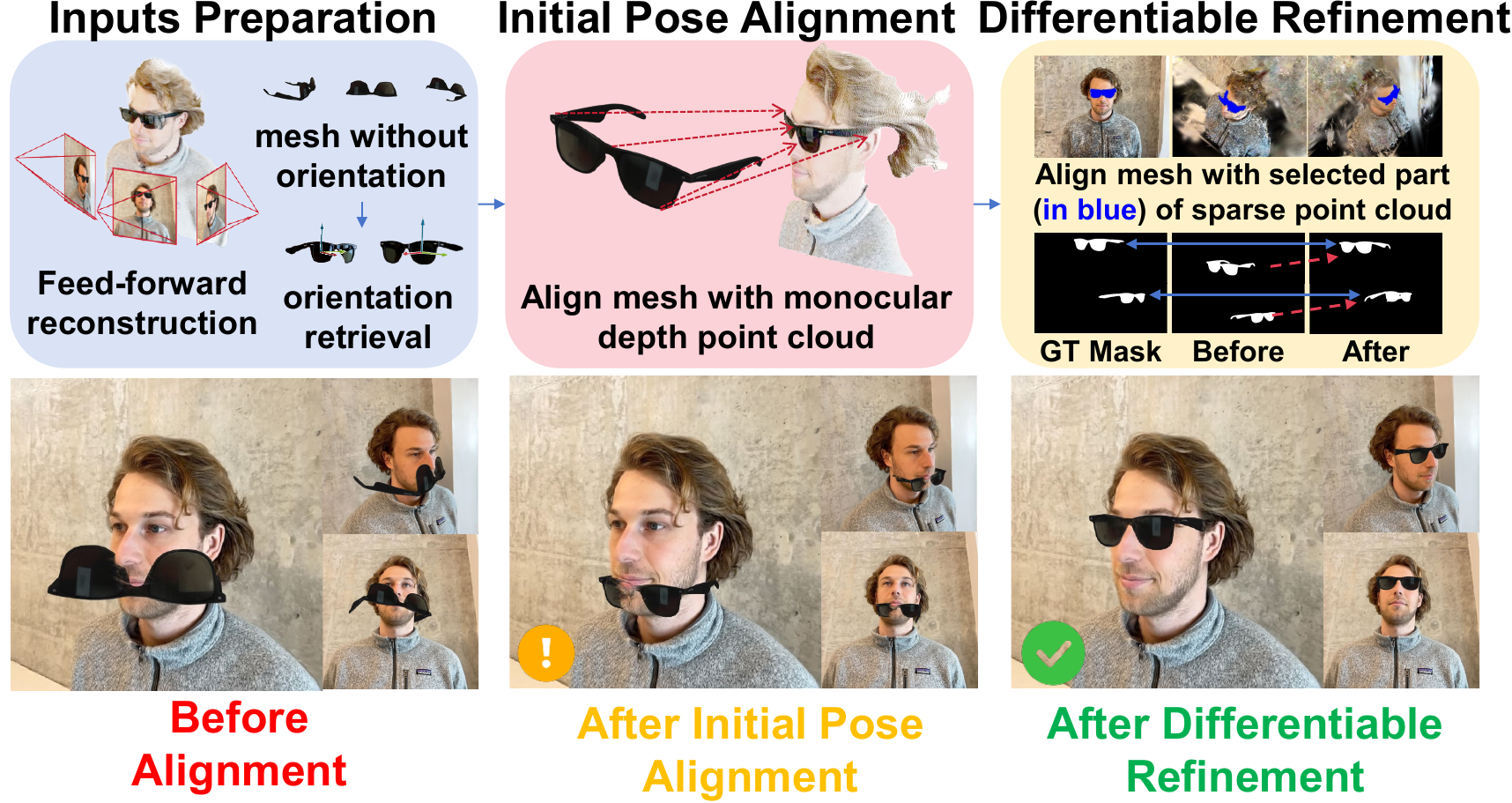}
    \caption{Two-stage TGA alignment. Phase 1 initializes orientation and a coarse similarity transform; Phase 2 first performs 7-DoF coarse docking and then fixes rotation for fine scale/translation anchoring. The bottom row is a representative case of adding sunglasses to the man’s face.}
    \label{fig:alignment_pipe}
\end{figure}
\paragraph{Progressive Coordinate Registration and Alignment.}
\label{subsec:two_stage_align}
We align generated assets to the 3DGS scene with two phases (see~\cref{fig:alignment_pipe}). \textbf{Phase 1: Initial Pose Alignment.} We render six canonical mesh views, estimate the up direction with orientation priors, and use dense render--scene matching to reject flipped or inverse-order correspondences. The 2D matches are lifted by the rendered mesh depth and monocular scene depth, giving paired source points $P_{\text{Mesh}}^i$ in the mesh coordinate system and target points $P_{\text{Mono}}^i$ in the scene coordinate system without manual point selection or rotation. We then solve for $(s,\mathbf{R},\mathbf{t})$ with Procrustes-style similarity alignment:
\begin{equation}\min_{s, \mathbf{R}, \mathbf{t}} \sum_i \|P_{\text{Mono}}^i - (s\mathbf{R}P_{\text{Mesh}}^i + \mathbf{t})\|_2^2.\end{equation}
This initializes $\mathbf{R}$ for differentiable registration; additional thresholds and failure cases are provided in the supplement.
\textbf{Phase 2: Differentiable Refinement.} With rotation initialized, we avoid flipped-pose local minima and align the mesh to a target sparse point cloud~\cite{wang2025vggt, lin2025depth} $\mathcal{S}_{\text{tar}}$, obtained from the scene or feed-forward reconstruction, using:
\begin{equation}
\begin{aligned}
\mathcal{L}_{\text{refine}} ={}& \lambda_{\text{geo}}\mathcal{L}_{\text{geo}}
+ \lambda_{\text{mask}}\mathcal{L}_{\text{mask}}
+ \mathbb{1}_{\mathrm{SDF}}(\lambda_{\text{sdf}}\mathcal{L}_{\text{SDF}}+\mathcal{R}_{\text{reg}}).
\end{aligned}
\end{equation}
where $\mathcal{L}_{\text{geo}} = \mathcal{L}_{\text{CD}} + \mathcal{L}_{\text{bbox}}$ combines Chamfer distance to $\mathcal{S}_{\text{tar}}$ with an axis-aligned bbox min/max MSE term, and $\mathcal{L}_{\text{mask}}$ is the MSE between rasterized mesh masks and multi-view target masks. Let $d(\cdot)$ be the local voxel SDF of $\mathcal{S}_{\text{tar}}$, positive outside the observed surface and negative behind it. Its penetration term is $\mathcal{L}_{\text{SDF}}=|\mathcal{X}|^{-1}\sum_{\mathbf{x}\in\mathcal{X}}\max(0,\tau-d(\mathbf{x}))^2$, where $\mathcal{X}$ are transformed mesh samples and $\tau=0.01$ is a safety margin for the local voxel SDF proxy. This one-sided hinge penalizes penetration and near-surface overlap caused by noise in the point-cloud-derived SDF while leaving samples beyond the margin unaffected. In Phase 2, coarse docking first optimizes the 7-DoF similarity transform, consisting of scale $s$, translation $\mathbf{t}$, and a normalized quaternion representing 3 rotational DoF, with $\mathbb{1}_{\mathrm{SDF}}=0$. Fine anchoring then enables $\mathbb{1}_{\mathrm{SDF}}=1$, fixes $\mathbf{R}$, and optimizes only $s$ and $\mathbf{t}$ with $\mathcal{R}_{\text{reg}}=\lambda_t\|\mathbf{t}-\mathbf{t}_0\|_2^2+\lambda_s(s-s_0)^2$, where $(s_0,\mathbf{t}_0)$ are the coarse-docking estimates. Since rotation is fixed in fine anchoring, no quaternion regularizer is used in this stage. Loss weights and schedule are provided in the supplement.

\subsection{Stage II: Appearance Harmonization}\label{subsec:CVM}
Direct mesh projection can introduce seams and miss local illumination, shadows, and reflections. CVM addresses them with \textbf{Adaptive Trajectory Synthesis} for $\rho$-controlled sampling and \textbf{Contextual Mask Refinement} for repainting projected foregrounds, boundaries, and nearby surfaces (see~\cref{fig:pipeline_overview}, bottom row). We instantiate $\Phi$ with Wan2.1; implementation details are in the supplement.

\begin{figure}[t]
    \centering
    \includegraphics[width=\columnwidth]{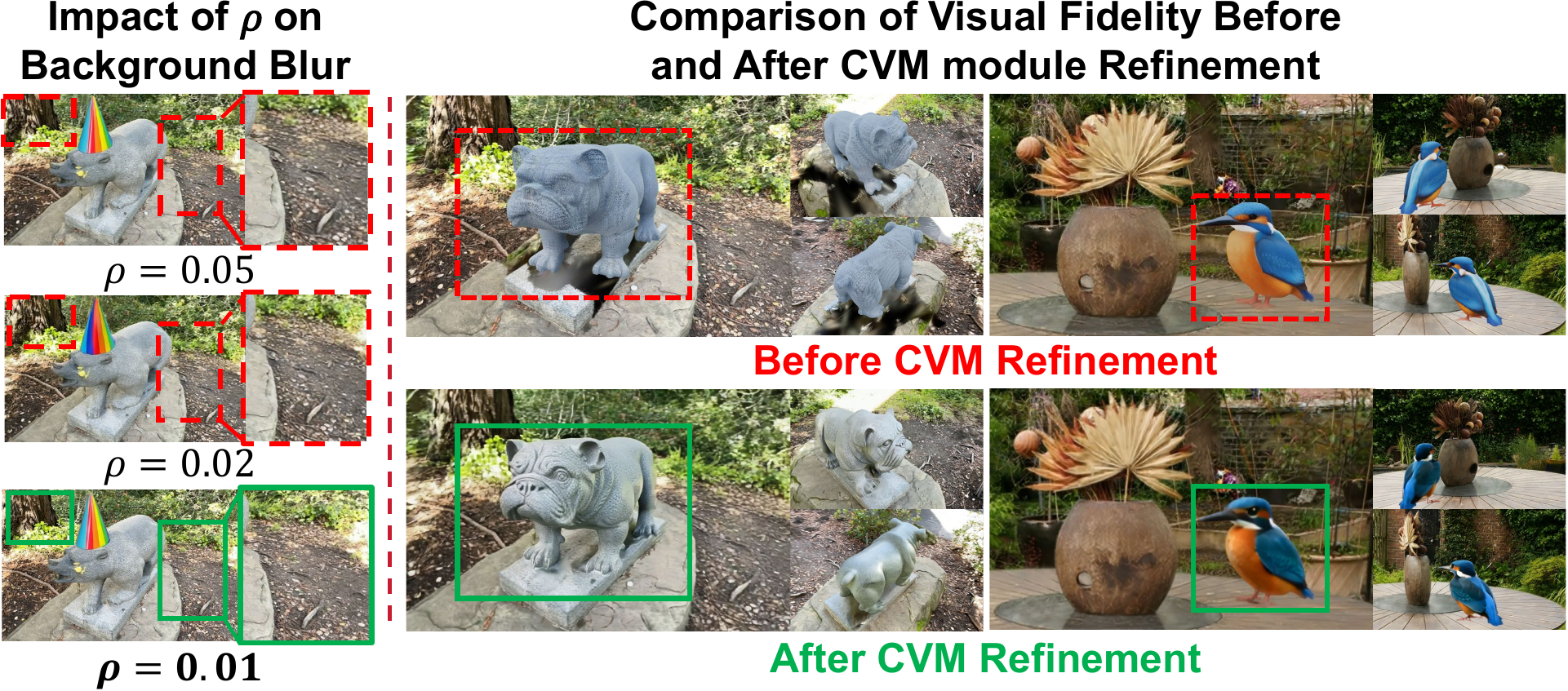}
    \caption{Selected CVM and trajectory examples. Left: frames at three trajectory densities $\rho$ (smaller is denser). Right: boundary crops before/after refinement.}
    \label{fig:texture_ablation}
\end{figure}

\subsubsection{Adaptive Trajectory Synthesis.} We select sparse key viewpoints from input camera distribution and connect them with SLERP-interpolated views to obtain smooth trajectories. For a segment with $N_{\text{seg}}$ frames, the sampling density is measured by:
$\rho = \frac{1}{N_{\text{seg}}-1} \sum_{t=1}^{N_{\text{seg}}-1} \| \log( \mathbf{r}_{t-1}^{-1} \mathbf{r}_{t} ) \|_2$
where $\mathbf{r}_t$ is the camera rotation at frame $t$; $\mathbf{r}_{t-1}^{-1} \mathbf{r}_{t}$ is the adjacent-frame increment. Keeping $\rho$ below a threshold preserves angular resolution and reduces motion blur (see Fig.~\ref{fig:texture_ablation}, left).
\subsubsection{Contextual Masking and Refinement}
To lift 2D modifications into 3D space, CVM transforms the rendered sequence $\mathcal{V}$ into an edited version $\mathcal{V}^*$. For each view, CVM partitions pixels into a geometry-defined core mask $\mathbf{M}^{core}_v$, an adaptive context mask $\mathbf{M}^{ctx}_v$, and a preserved background $\mathbf{M}^{keep}_v$. The core mask comes from mesh projection, removal regions, disocclusions, or the union of old/new projections for replacement. The context mask is initialized from the self-refinement difference between the first frame of $\mathcal{V}$ and its locally repaired version. The resulting region is propagated along the rendered video and locally dilated to cover object boundaries and nearby appearance changes, such as contact shadows and reflections. Thus $\mathbf{M}^{edit}_v=\mathbf{M}^{core}_v\cup\mathbf{M}^{ctx}_v$ and $\mathbf{M}^{keep}_v=1-\mathbf{M}^{edit}_v$. For long-duration edits, CVM employs an autoregressive strategy across overlapping segments $\{S_1, \dots, S_m\}$. In segment $S_j$, the Wan2.1 generator $\Phi$ maps the input latent sequence $\mathbf{z}_j$ to an edited latent sequence $\mathbf{z}_j^*$ under the text prompt, editable mask, gray-prefilled context, and decoded overlap frames from the previous segment:
\begin{equation}\mathbf{z}_j^* = \Phi \left( \mathbf{z}_j \mid y, \mathbf{M}^{edit}_j, \mathcal{V}_j^{\text{gray}}, \widehat{\mathcal{V}}_{j-1}^{\text{tail}} \right).
\end{equation}
Here, $\mathcal{V}_j^{\text{gray}}$ denotes the conditioning frames in which editable pixels are gray-filled and preserved pixels remain visible, and $\widehat{\mathcal{V}}_{j-1}^{\text{tail}}$ denotes the decoded edited frames from the tail of segment $S_{j-1}$. Let $I^{CVM}_v$ be the decoded CVM output frame and $I^{src}_v$ the original rendered frame for view $v$. The final frame is composited as $I^{out}_v=\mathbf{M}^{edit}_v I^{CVM}_v+\mathbf{M}^{keep}_v I^{src}_v$, preserving the background (see Fig.~\ref{fig:texture_ablation}, right). Finally, $\mathcal{V}^*$ reconstructs $\mathcal{G}^*$ from $\mathcal{G}$ under $\mathcal{C}$ using mask-gated photometric gradients without mesh loss using splatfacto.

%% file: sec/4_experiment.tex
\begin{figure*}[t!]
  \centering
  \includegraphics[width=\textwidth]{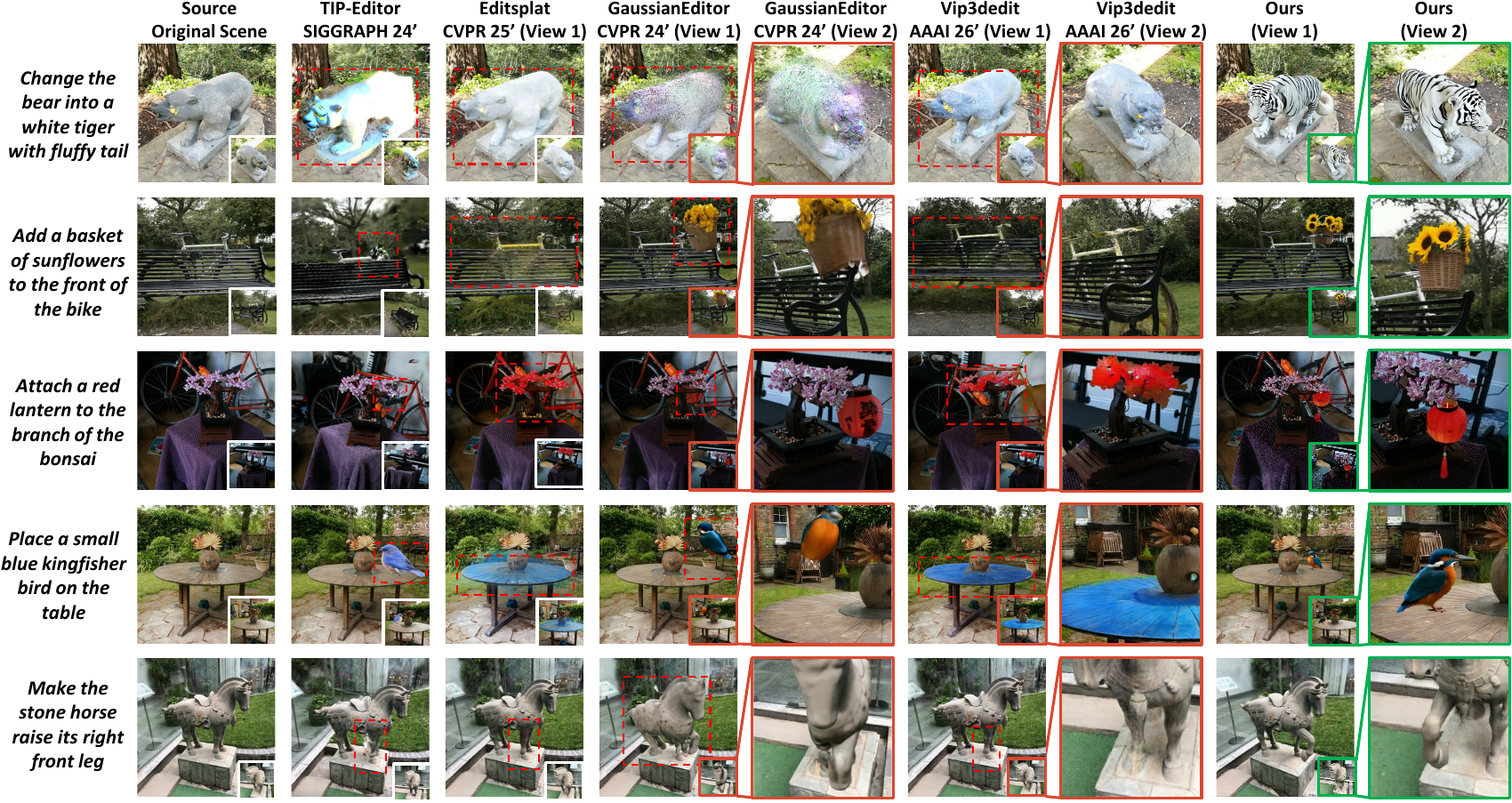}
  \caption{Qualitative comparisons. \textcolor{red}{Red} boxes highlight baseline artifacts, while \textcolor[RGB]{0,128,0}{green} boxes show TRACE's more consistent geometry and cleaner local details. For clarity, we enlarge the second-view results of TRACE and selected competitive baselines.}
  \label{fig:qualitative_comp_1}
\end{figure*}

\input{sec/4_main_table}

\section{Experiments}
\label{sec:experiments}

\subsection{Experimental Setup}
Following GaussianEditor~\cite{chen2023gaussianeditor} and EditSplat~\cite{lee2025editsplat}, we evaluate on eight held-out scenes: 4 IN2N~\cite{instructnerf2023}, 1 BlendedMVS~\cite{yao2020blendedmvs}, and 3 Mip-NeRF 360~\cite{barron2022mip}. We evaluate local and global style change, insertion, removal, replacement, and deformation; the first two form \textbf{Global/Local Style Editing} and the remaining four \textbf{Partial/Whole Object Geometry Editing} in~\cref{tab:quantitative_results}. Training and evaluation scenes are disjoint. We report CLIP~\cite{radford2021learning} Directional Similarity and CLIP Similarity for semantic alignment, DINO Similarity~\cite{oquab2023dinov2} as a rendered-view appearance-consistency proxy, and aesthetic score~\cite{schuhmann2022laion5bopenlargescaledataset} for visual quality. TRACE takes approximately 10 minutes on one NVIDIA RTX Pro6000. Baselines use official settings, prompts, and resolution; runtimes include reconstruction but exclude model loading.

\subsection{Comparative Experiments}
\textbf{Qualitative Results.} As shown in~\cref{fig:qualitative_comp_1}, TRACE better preserves geometry and texture across deformation, insertion, and style changes, maintaining spatially consistent inserted objects and cleaner local details than other methods. Full-trajectory 360-degree renderings for the evaluation scenes are provided in the supplement.

\input{sec/4_user_study_figure}

\noindent\textbf{Quantitative Results.} As shown in~\cref{tab:quantitative_results}, TRACE achieves the best CLIP$_{dir}$, CLIP$_{sim}$, and aesthetic scores across both task groups under the common-prompt protocol. For geometry edits, we further report MEt3R~\cite{asim25met3r} and Uni3D-T~\cite{zhou2023uni3d} on final reconstructed regions: MEt3R directly measures multi-view 3D consistency, while Uni3D-T measures semantic similarity between the edited-region point cloud and edit text. As shown in~\cref{fig:user_study_summary}, fifteen annotators completed 36 randomized blind comparisons each against GaussianEditor, EditSplat, and Vip3dedit (540 comparison units), rating four criteria averaged into an overall preference: instruction following, visual quality, multi-view consistency, and edit locality/content preservation. Weighted scoring details are provided in the appendix.

\subsubsection{Qualitative Comparison with Direct Video Editing Methods}
\begin{figure}[t]
    \centering
    \includegraphics[width=\columnwidth]{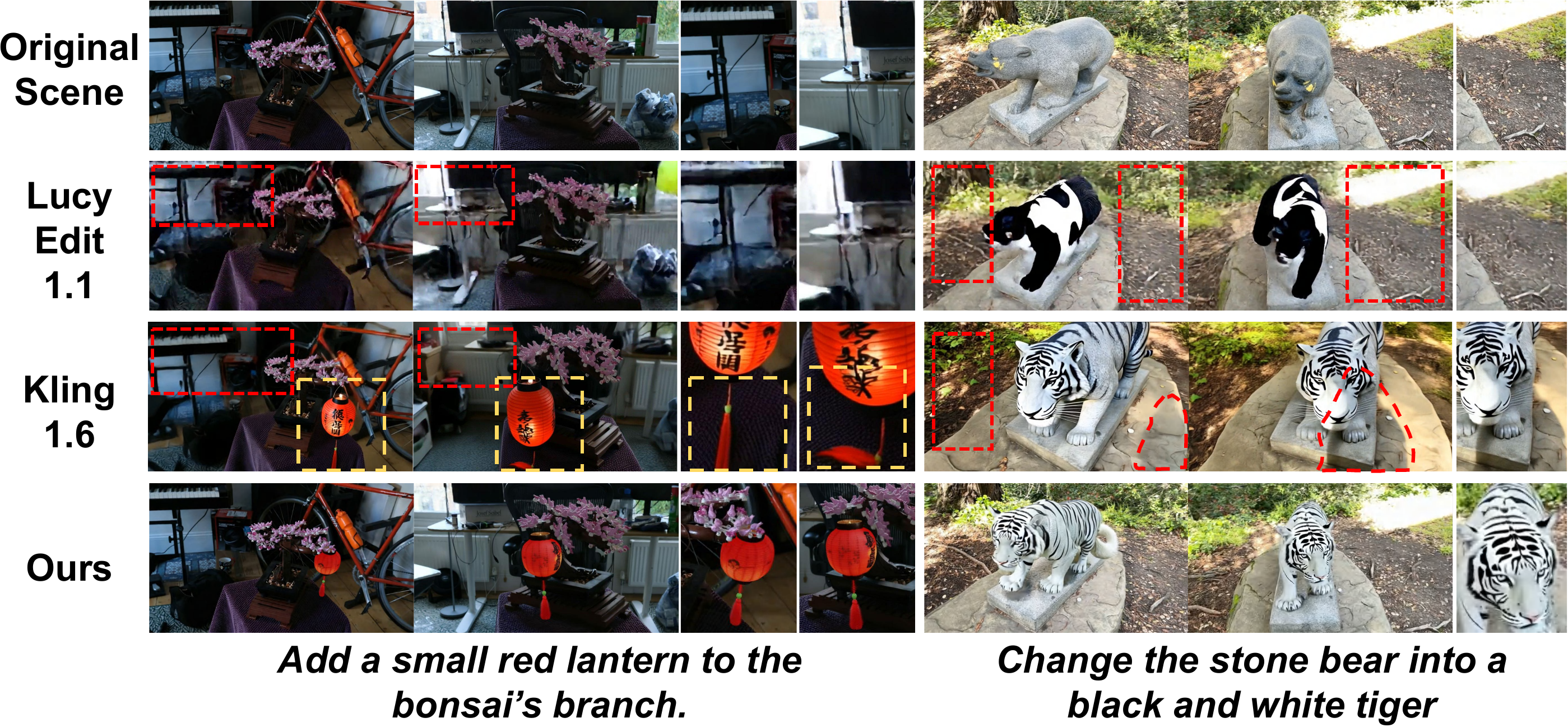}
    \caption{Comparison with direct video editing. \textcolor[RGB]{255,191,0}{Yellow} marks subject synthesis and \textcolor{red}{red} marks background stability; TRACE better preserves 3D placement and scene details.}
    \label{fig:video_editing_comp}
\end{figure}

As shown in~\cref{fig:video_editing_comp}, compared with Lucy Edit 1.1~\cite{team2025lucy} and Kling 1.6~\cite{kling}, TRACE avoids misalignment, shape-shifting, and background distortion.

\input{sec/4_ablation_table}

\subsection{Ablation and Analysis}
We ablate 3D-Anchor synthesis, TGA, and CVM, which provide sparse-view anchors, mesh alignment, and context-aware harmonization.
\begin{figure}[t]
    \centering
    \includegraphics[width=\columnwidth]{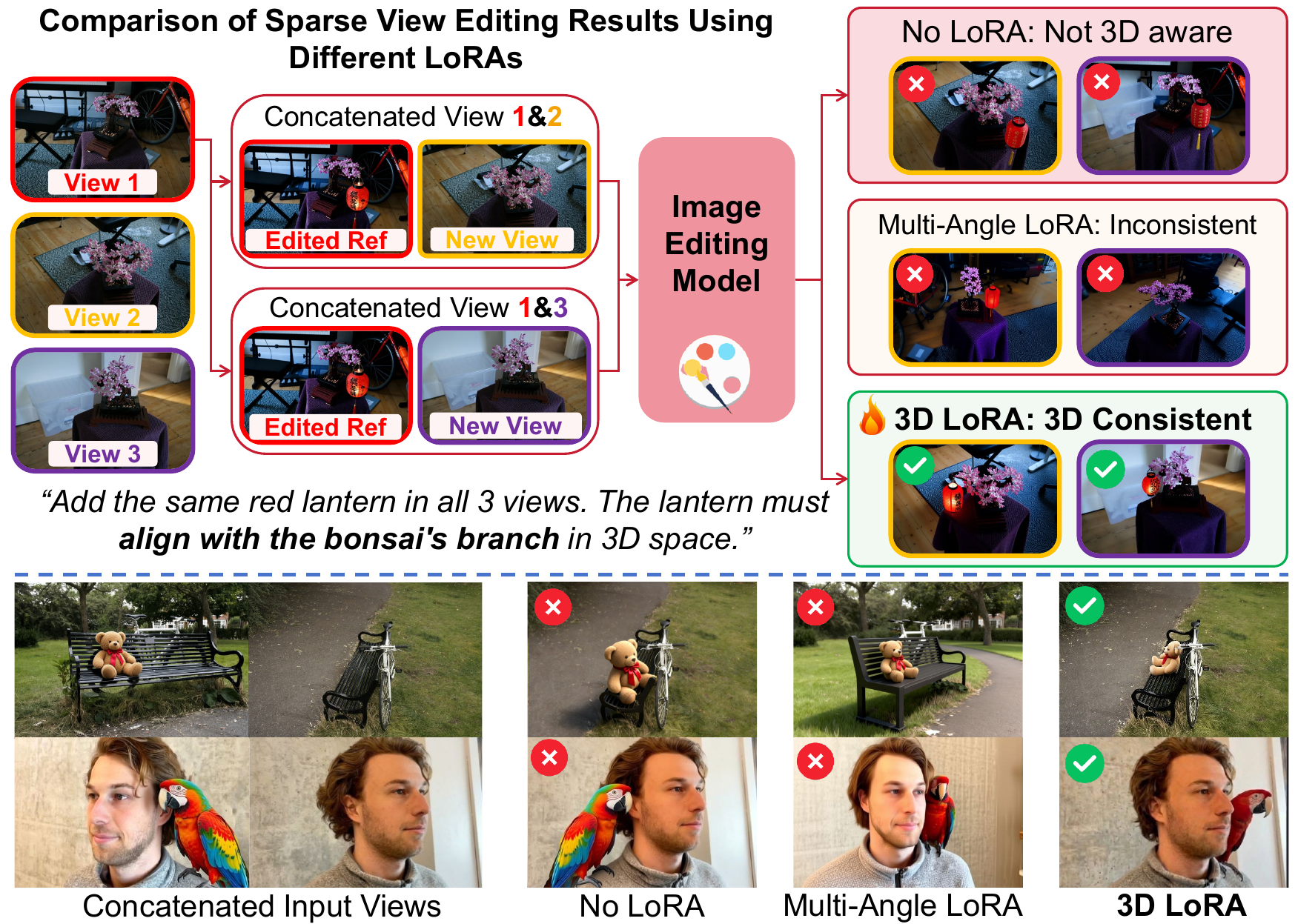}
    \caption{Multi-view editing ablation. View 1/2/3 are marked in \textcolor{red}{red}, \textcolor[RGB]{255,191,0}{yellow}, and \textcolor{violet}{purple}. No LoRA lacks 3D awareness, Multi-Angle LoRA suffers from inconsistent placement and background collapse, while 3D-LoRA preserves both object placement and scene background.}
    \label{fig:multiview_ablation}
\end{figure}

\subsubsection{Effectiveness of 3D-LoRA in multi-view anchor synthesis.}
We compare 3D-LoRA with No-LoRA iterative editing, Multi-Angle LoRA~\cite{qwen_image_edit_lora_2025}, and Concurrent Triple-view synthesis. As shown in Table~\ref{tab:ablation}(a) and~\cref{fig:multiview_ablation}, the baselines suffer from weak 3D awareness, fixed-angle placement errors, or cross-view interference. By learning scene-coherent placement from MV-TRACE, our 3D-LoRA preserves both object location and background fidelity, achieving the best $\mathrm{PSNR_{bg}}$, lowest $\mathrm{L_{LPIPS}}$, and highest $\mathrm{IoU_{align}}$.

\subsubsection{Ablation on CVM Refinement.}
The CVM variants in Table~\ref{tab:ablation}(c) evaluate semantic alignment, rendered-view similarity, and visual quality. Removing autoregressive propagation produces the largest DINO decrease, while removing contextual masks lowers CLIP and Aesthetic scores. Full CVM has nearly the same DINO value as w/o CVM ($0.9058$ vs. $0.9052$) but higher CLIP and Aesthetic scores.

\subsubsection{Ablation on TGA Module's Two-Phase Registration.}
\label{subsec:alignment_ablation}
We evaluate TGA using Avg-$IoU_{2D}$, $IoU_{3D}$, and the fraction of cases exceeding $IoU_{3D}$ thresholds of 0.50 and 0.80. The 0.50 threshold indicates sufficient overlap for successful alignment, whereas 0.80 denotes high-accuracy alignment. FreeInsert~\cite{li2025freeinsert} is included only as an external mesh-scene alignment reference, since it is a specialized insertion method with VLM grounding and placement steps. As shown in Table~\ref{tab:ablation}(b), full TGA is close to FreeInsert in average 2D/3D IoU and obtains higher threshold success rates, indicating robust alignment quality. Removing alignment nearly fails, while the two phases complement coarse orientation estimation with scale and translation refinement.
\FloatBarrier

%% file: sec/4_main_table.tex
\begin{table*}[t!]
  \centering
  \footnotesize
  \setlength{\tabcolsep}{0pt}
  \setlength{\aboverulesep}{0.25ex}
  \setlength{\belowrulesep}{0.25ex}
  \renewcommand{\arraystretch}{1.08}
  \begin{tabular*}{\textwidth}{@{\extracolsep{\fill}}lcccccccccccc@{}}
    \toprule
    & & \multicolumn{4}{c}{\textbf{Global/Local Style Editing}}
    & \multicolumn{6}{c}{\textbf{Partial/Whole Object Geometry Editing}}
    & \multicolumn{1}{c}{\textbf{Time}} \\
    \cmidrule(lr){3-6}\cmidrule(lr){7-12}\cmidrule(lr){13-13}
    \textbf{Method}
    & \textbf{Pub.}
    & CLIP$_{dir}\uparrow$ & CLIP$_{sim}\uparrow$ & DINO$\uparrow$ & Aes.$\uparrow$
    & CLIP$_{dir}\uparrow$ & CLIP$_{sim}\uparrow$ & DINO$\uparrow$ & Aes.$\uparrow$ & MEt3R$\downarrow$ & Uni3D-T$\uparrow$
    & min$\downarrow$ \\
    \midrule
    DGE & ECCV'24 & 0.1082 & 0.2337 & 0.8693 & 5.5453 & 0.0228 & 0.2405 & 0.9005 & \second{6.0495} & 0.2095 & 0.0574 & 10 min\\
    GaussCtrl & ECCV'24 & 0.0349 & 0.1955 & \second{0.8962} & 5.3021 & 0.0543 & 0.2297 & 0.8962 & 5.7307 & 0.1820 & 0.0042 & 20 min\\
    TIP-Editor & SIG'24 & \second{0.1095} & \second{0.2398} & 0.8693 & 5.2792 & \second{0.0931} & 0.2234 & 0.8375 & 5.4916 & 0.2318 & 0.1531 & 45 min\\
    GaussianEditor & CVPR'24 & 0.0737 & 0.2136 & 0.8774 & 5.6241 & 0.0623 & \second{0.2372} & 0.8568 & 5.2373 & 0.1847 & \second{0.1690} & 16 min\\
    EditSplat & CVPR'25 & 0.0790 & 0.2328 & 0.8658 & 5.7483 & 0.0734 & 0.2270 & \second{0.9010} & 5.8659 & \second{0.1645} & 0.0283 & 18 min\\
    Vip3dedit & AAAI'26 & 0.0189 & 0.2172 & 0.8806 & \second{5.8072} & 0.0473 & 0.2136 & 0.8856 & 5.5574 & 0.2079 & 0.0350 & 10 min\\
    \oursrow \textbf{TRACE} & \textbf{---} & \best{0.1147} & \best{0.2410} & \best{0.9092} & \best{6.0638} & \best{0.1881} & \best{0.2520} & \best{0.9024} & \best{6.1432} & \best{0.1521} & \best{0.2013} & \best{10 min} \\
    \bottomrule
  \end{tabular*}
  \caption{Unified evaluation on style and geometry editing. The evaluation covers local style change, global style change, insertion, removal, replacement, and deformation. MEt3R and Uni3D-T are reported for final reconstructed geometry-editing regions.}
  \label{tab:quantitative_results}
\end{table*}

%% file: sec/4_user_study_figure.tex
\begin{figure}[b!]
  \centering
  \resizebox{\columnwidth}{!}{\input{Figures/user_study_pairwise}}
  \captionsetup{font=normalsize}
  \caption{Pairwise user preference across six editing types.
  Results aggregate 540 blind comparison units as normalized weighted preference strengths. Tasks 1--6 denote local/global style, insertion, removal, replacement, and deformation.}
  \label{fig:user_study_summary}
\end{figure}

%% file: Figures/user_study_pairwise.tex
\begingroup
\definecolor{userOurs}{HTML}{7FBEA8}
\definecolor{userTie}{HTML}{B9B9B9}
\definecolor{userBaseline}{HTML}{EE956B}

\newcommand{\studybar}[5]{%
  \pgfmathsetmacro{\studyOne}{#2+#3}%
  \pgfmathsetmacro{\studyTwo}{#2+#3+#4}%
  \pgfmathsetmacro{\studyEnd}{#2+#3+#4+#5}%
  \fill[userOurs] (#1,#2) rectangle ({#1+4.25},\studyOne);
  \fill[userTie] (#1,\studyOne) rectangle ({#1+4.25},\studyTwo);
  \fill[userBaseline] (#1,\studyTwo) rectangle ({#1+4.25},\studyEnd);
  \node[font=\sffamily\scriptsize\bfseries,text=black!80]
    at ({#1+2.125},{#2+#3/2}) {#3};
  \node[font=\sffamily\scriptsize\bfseries,text=black!80]
    at ({#1+2.125},{#2+#3+#4/2}) {#4};
  \node[font=\sffamily\scriptsize\bfseries,text=black!80]
    at ({#1+2.125},{#2+#3+#4+#5/2}) {#5};
}

\newcommand{\studygrid}[2]{%
  \foreach \y in {20,40,60,80,100}{
    \draw[black!10,line width=0.3pt] (#1,\y)--({#1+32.75},\y);
  }
  \draw[black!45,line width=0.4pt] (#1,0) rectangle ({#1+32.75},100);
  \node[font=\sffamily\Large\bfseries] at ({#1+16.375},108) {#2};
}

\newcommand{\studylabels}[1]{%
  \node[draw=black!45,fill=black!4,circle,inner sep=0.4pt,minimum size=8.5pt,
        font=\sffamily\Large\bfseries,text=black!80] at ({#1+2.125},-6) {1};
  \node[draw=black!45,fill=black!4,circle,inner sep=0.4pt,minimum size=8.5pt,
        font=\sffamily\Large\bfseries,text=black!80] at ({#1+7.425},-6) {2};
  \node[draw=black!45,fill=black!4,circle,inner sep=0.4pt,minimum size=8.5pt,
        font=\sffamily\Large\bfseries,text=black!80] at ({#1+12.725},-6) {3};
  \node[draw=black!45,fill=black!4,circle,inner sep=0.4pt,minimum size=8.5pt,
        font=\sffamily\Large\bfseries,text=black!80] at ({#1+18.025},-6) {4};
  \node[draw=black!45,fill=black!4,circle,inner sep=0.4pt,minimum size=8.5pt,
        font=\sffamily\Large\bfseries,text=black!80] at ({#1+23.325},-6) {5};
  \node[draw=black!45,fill=black!4,circle,inner sep=0.4pt,minimum size=8.5pt,
        font=\sffamily\Large\bfseries,text=black!80] at ({#1+28.625},-6) {6};
}

\begin{tikzpicture}[x=0.128cm,y=0.034cm]
  \path[use as bounding box] (-11,-44) rectangle (115,113);

  \studygrid{0}{GaussianEditor}
  \studygrid{41}{EditSplat}
  \studygrid{82}{Vip3dedit}

  \foreach \y in {0,20,40,60,80,100}{
    \node[font=\sffamily\Large,anchor=east] at (-1.1,\y) {\y};
  }
  \node[font=\sffamily\Large\bfseries,rotate=90] at (-9.0,50) {Weighted (\%)};

  \studybar{1.0}{0}{56.9}{17.2}{25.9}
  \studybar{6.3}{0}{58.0}{20.0}{22.0}
  \studybar{11.6}{0}{76.0}{12.0}{12.0}
  \studybar{16.9}{0}{78.0}{10.2}{11.9}
  \studybar{22.2}{0}{74.0}{16.0}{10.0}
  \studybar{27.5}{0}{72.0}{16.0}{12.0}

  \studybar{42.0}{0}{56.2}{16.7}{27.1}
  \studybar{47.3}{0}{59.0}{15.4}{25.6}
  \studybar{52.6}{0}{64.0}{20.0}{16.0}
  \studybar{57.9}{0}{62.0}{20.0}{18.0}
  \studybar{63.2}{0}{60.0}{20.0}{20.0}
  \studybar{68.5}{0}{66.0}{18.9}{15.1}

  \studybar{83.0}{0}{57.5}{20.0}{22.5}
  \studybar{88.3}{0}{56.5}{17.4}{26.1}
  \studybar{93.6}{0}{70.0}{20.0}{10.0}
  \studybar{98.9}{0}{68.0}{20.0}{12.0}
  \studybar{104.2}{0}{56.0}{20.0}{24.0}
  \studybar{109.5}{0}{56.0}{16.0}{28.0}

  \studylabels{0}
  \studylabels{41}
  \studylabels{82}

  \fill[userOurs] (3,-44) rectangle (8,-36);
  \node[font=\sffamily\Large\bfseries,anchor=west] at (10,-40) {TRACE Preferred};
  \fill[userTie] (41,-44) rectangle (46,-36);
  \node[font=\sffamily\Large\bfseries,anchor=west] at (48,-40) {No Preference};
  \fill[userBaseline] (74,-44) rectangle (79,-36);
  \node[font=\sffamily\Large\bfseries,anchor=west] at (81,-40) {Baseline Preferred};
\end{tikzpicture}
\endgroup

%% file: sec/4_ablation_table.tex
\begin{table}[b!]
    \centering
    \setlength{\tabcolsep}{1.2pt}
    \setlength{\aboverulesep}{0.25ex}
    \setlength{\belowrulesep}{0.25ex}

    {\footnotesize
    \renewcommand{\arraystretch}{1.02}
    \begin{tabularx}{.95\columnwidth}{@{}lYYY@{}}
      \multicolumn{4}{c}{(a) Ablation on 3D-Anchor Synthesis} \\
      \toprule
      Method & $L_{\text{LPIPS}}\downarrow$ & $\text{PSNR}_{\text{bg}}\uparrow$ & $IoU_{\text{align}}\uparrow$ \\
      \midrule
      No LoRA & 0.224 & 26.8 & 0.190 \\
      Multi-Angle LoRA & 0.205 & \second{28.5} & \second{0.588} \\
      Triplet & \second{0.198} & 24.2 & 0.240 \\
      \oursrow \textbf{Ours (3D-LoRA)} & \best{0.144} & \best{33.1} & \best{0.815} \\
      \bottomrule
    \end{tabularx}

    \begin{tabularx}{.98\columnwidth}{@{}lYYYY@{}}
      \multicolumn{5}{c}{(b) Ablation on TGA Module} \\
      \toprule
      Config & $IoU_{2D}\uparrow$ & $IoU_{3D}\uparrow$ & \makecell[c]{Align. Acc.\\@0.50$\uparrow$} & \makecell[c]{Align. Acc.\\@0.80$\uparrow$} \\
      \midrule
      w/o Align & 0.012 & 0.000 & 0.0\% & 0.0\% \\
      Phase 1 only & 0.213 & 0.031 & 16.7\% & 8.3\% \\
      Phase 2 only & 0.538 & 0.635 & 54.2\% & 37.5\% \\
      FreeInsert & \second{0.847} & \second{0.803} & \second{83.3\%} & \second{70.8\%} \\
      \oursrow \textbf{Ours} & \best{0.852} & \best{0.815} & \best{91.7\%} & \best{79.1\%} \\
      \bottomrule
    \end{tabularx}

    \begin{tabularx}{.95\columnwidth}{@{}lYYYY@{}}
      \multicolumn{5}{c}{(c) Ablation on CVM Module} \\
      \toprule
      Method & \makecell[c]{CLIP\\Dir.$\uparrow$} & \makecell[c]{CLIP\\Sim.$\uparrow$} & \makecell[c]{DINO\\Sim.$\uparrow$} & \makecell[c]{Aes.\\Score$\uparrow$} \\
      \midrule
      w/o CVM & 0.1254 & 0.2343 & \second{0.9052} & 5.4810 \\
      w/o AR & \second{0.1485} & 0.2388 & 0.8524 & \second{5.9439} \\
      w/o Ctx. & 0.1412 & \second{0.2405} & 0.8930 & 5.8821 \\
      \oursrow \textbf{Ours} & \best{0.1514} & \best{0.2465} & \best{0.9058} & \best{6.1035} \\
      \bottomrule
    \end{tabularx}
    }
    \caption{Ablation studies of TRACE's key components: 3D-Anchor synthesis, TGA alignment, and CVM refinement.}
    \label{tab:ablation}
\end{table}

%% file: sec/5_conclusion.tex


\section{Conclusion}
\label{sec:conclusion}
We present \textbf{TRACE}, a mesh-guided 3DGS editing framework that decouples \textbf{Geometric Anchoring} from \textbf{Appearance Harmonization}. Multi-view 3D-Anchor Synthesis and coarse-to-fine TGA establish scene-aligned geometry, while autoregressive CVM harmonizes projected anchors across views, enabling diverse and structurally coherent edits with improved structural integrity and scene coherence.